\documentclass[twocolumn]{article}

\usepackage{arxiv}

\usepackage[utf8]{inputenc}
\usepackage[T1]{fontenc}
\usepackage[colorlinks=true,linkcolor=blue,citecolor=blue,urlcolor=blue]{hyperref}
\usepackage{url}
\usepackage{booktabs}
\usepackage{amsmath,amssymb,amsfonts}
\usepackage{amsthm}
\usepackage{multirow}
\usepackage{microtype}
\usepackage{graphicx}
\usepackage[authoryear]{natbib}
\setcitestyle{round,semicolon}
\usepackage{doi}
\usepackage{nicefrac}
\usepackage{capt-of}
\usepackage{cleveref}

\theoremstyle{remark}
\newtheorem{remark}{Remark}

\title{Why shared attention vectors fail: a case for outcome-indexed tuning}

\date{}

\author{%
	\href{https://orcid.org/0000-0001-7487-1974}{\includegraphics[scale=0.06]{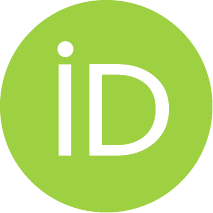}\hspace{1mm}Lenard~Dome} \\
	Department of Psychiatry and Psychotherapy, Faculty of Medicine\\
	University of T\"ubingen, T\"ubingen, Germany\\
	German Center for Mental Health (DZPG), T\"ubingen, Germany\\
	\texttt{lenard.dome@uni-tuebingen.de} \\
}

\renewcommand{\shorttitle}{Why shared attention vectors fail}

\hypersetup{
pdftitle={Why shared attention vectors fail: a case for outcome-indexed tuning},
pdfsubject={q-bio.NC},
pdfauthor={Lenard Dome},
pdfkeywords={outcome-indexed attention, feed-forward networks, attention shift, multi-outcome learning, gradient descent on error},
}

\begin{document}
\twocolumn[{
  \begin{@twocolumnfalse}

	\maketitle

	\begin{abstract}
		Dimensional attention in learning is often implemented as a globally shared attention vector, where each stimulus dimension corresponds to a single scalar. These scalars are learned by models through gradient-descent on error, where predictive features acquire more salience. We show that under multi-outcome learning, where models predict more than one outcome, this shared vector becomes unstable; it collapses to its bounds and prevents the models from learning meaningful attentional tunings for learning and generalization. We address this by introducing an outcome-indexed attentional matrix that converts globally shared attentional tuning into an outcome-indexed representation. We present an analysis of the unstable shared vectors and derive the conditions under which it holds. Empirically, three synthetic experiments benchmark the proposed attention matrices and show that they converge to meaningful representations, something shared attention vectors fail to do. These results suggest that outcome-indexed attentional matrices are a general fix for gradient-based attentional processes, which improves models of learning under multi-outcome conditions.
	\end{abstract}

  \end{@twocolumnfalse}
  \vspace{1em}
}]

\section{Introduction}\label{sec:introduction}

Attention is a quintessential part of learning. The capacity of an organism--or any information-processing system--to choose and direct itself towards diagnostic information, while disregarding irrelevant ones, is a powerful adaptive ability. As a driving explanatory mechanism, attention has been shown to account for a wide range of phenomena \citep{lepelley2016attention,don2021hearing, paskewitz2020dissecting, paskewitz2023statistical, livesey2025attention}. Selective attention to features of a stimulus has also been proposed as a formal requirement for models of categorization \citep{kruschke1993three}.

One of the most influential attentional frameworks was proposed by \cite{mackintosh1975theory}, who conceptualized attention as salience underlying each stimulus feature, represented by a feature-specific scalar -- collected in a vector of saliences. These saliences are derived from experience, where attention to a given feature is determined by how well that feature predicts a given outcome. Eye-tracking data extensively corroborated this attentional allocation account, where eye-fixation proportions were taken to correspond to the mechanisms specified in attentional theories \citep[e.g.][]{lepelley2016attention,easdale2019onset,beesley2015uncertainty,don2019learned,wills2007predictive,walker2019role,stojic2020its}. This framework has influenced a number of formal models \citep[e.g.][]{kruschke2001unified,kruschke1992alcove,paskewitz2020dissecting,paskewitz2023statistical}, which provide extensive formalism for the process of allocating attention to predictive features. In all of these instantiations, salience is feature-specific, clamped to $[0, \infty)$, stored in a globally shared attention vector, and adjusted using gradient descent on error.

In a recent work, we have discovered that globally shared attention vectors become volatile under specific parameter and environmental constraints \citep{dome2025gdistance}. Because attention weights are clamped between 0 and $\infty$, models reset attention weights to 0 every time the update pushes values below this boundary. This clamping mechanism is part of the model specification that discards genuinely non-diagnostic cues: the resetting of attentional weights is an intended consequence of large updates \citep{kruschke2001unified, paskewitz2020dissecting}. This resetting deploys a legitimate function in single-outcome cases. Here, we show that this mechanism becomes unstable under multi-outcome learning, remains independent of the step-size, and prevents models to learn what feature to attend. We identify the summation-across-outcomes as the primary cause, which makes the collapse of scalar salience values indiscriminate with respect to how predictive those features actually are (a feature diagnostic for outcome A still nets a below-boundary displacement once gradients from outcome B, C, are summed onto the globally shared salience scalar). We provide a principled solution that enables the encoding of a much richer attentional mapping. Furthermore, our solution improves model's capacity to account for complex real-world behavior.

\section{Preliminaries}\label{sec:preliminaries}

We consider the problem for a set of learning models that adjust salience weights to minimize error. These models use a nominal input representation that encodes presence or absence of a feature as 0 or 1, such that each stimulus is represented as vector $S = \{s_1, s_2, \dots, s_n\}$, with $n$ representing the number of possible features of which the stimulus can take. Following in the \cite{mackintosh1975theory} framework, each input to the system correspond to an underlying scalar salience, $\eta = \{\eta_1, \eta_2, \dots, \eta_n\}$ with a $[0, \infty)$ bound. The model then combines $S$ with $\eta$ to generate attention gains, $g$, at the beginning of each trial:

\begin{equation}
    g_i = \eta_i s_i
\end{equation}

These attention gains are normalized through their vector $p$-norm:

\begin{equation}
    a_i = \frac{g_i}{\left(\sum^{n}_{j=1} |g_j|^p\right)^{\frac{1}{p}}}
\end{equation}

with $p$ representing a brutality parameter, controlling the degree of attentional competition between currently present input dimensions. Most often, these normalized attention strengths are combined with connection weights to produce model predictions, $O$, along $K$ outcomes (e.g. usually implemented as output units in feed-forward connectionist networks):

\begin{equation}
    o_k = \sum_i w_{ki}a_i
\end{equation}

After making a prediction, models receive feedback. That feedback is used to calculate the following error for the derivations:

\begin{align}
    \delta_k &= \lambda_k - o_k, \\
    L &= \frac{1}{2}\sum_k{\delta_k^2}
\end{align}

where $\lambda$ is a teaching vector supplied to the model. Salience is adjusted via gradient descent on error: predictive stimuli acquires higher attention weights, whereas unpredictive stimuli reduces in salience. With the salience weights constrained to be non-negative via a hard projection (clamp), the attention shifts can be written as:

\begin{align}\label{eq:kruschke-shift}
\Delta g'_{i,j+1} &= - \rho \frac{\partial L}{\partial g_{i,j}}\text{,\,\,\; reiterates ten times} \\
g'_i &= \max\Big[0, g'_{j} \Big]
\end{align}

where $\rho$ is the step-size, $t$ is the trial, $j$ is the current iteration for the descent, and $g'$ is the updated attention gain for each iteration. Attention shifts are a non-linear function (gradient changes as attention changes), which means that the shift cannot be achieved with a single large step along the gradient. Therefore, Equation \ref{eq:kruschke-shift} reiterates 10 times. The starting value of $g'_{j=1}$ is the current $g$ on that particular trial. The expanded equation for Equation \ref{eq:kruschke-shift} for the attention shift yields

\begin{align}
\Delta g'_{j+1}
&= \rho s_i \|{g_j}'\|_p^{-1}\sum_k(W_{ki}s_i - a'^{p-1}_{i}o'_k)\delta'_k
\end{align}

where $o'$, $a'$, $g'$ are recalculated on each $j$ iteration. Critically, on any of these iterations, $g'$ values are clamped between $[0, +\infty)$. Finally, $\eta$ is updated via state displacement, similar to reconstruction error in recirculation networks \citep{hinton1987learning,oreilly1996biologically}, where $\alpha$ is a learning rate:

\begin{align}
\eta_{t+1} &= \max\!\Big[0,\; \eta_t + \alpha \,\left(g' - g\right)\Big].
\end{align}

\section{Problem Statement}\label{section:problem}

Scalar salience values intended to live between $[0,\infty)$ become unstable when aggressive updates force the salience to reset at $0$. The attention shifts (settling dynamics) operate on a single shared state (vector) $g$ with the aggregated loss:

\begin{equation}
L = \sum_{k} L_k
\end{equation}

where $L_k$ is the $k$-outcome loss: $\frac{1}{2}\delta_k^2$. The gradient descent used in the ten iterations of the attention shift sums over all outcome-specific error signal:

\begin{equation}\label{eq:sum-over-k}
\frac{\partial L}{\partial g} = \sum_{k} \frac{\partial L_k}{\partial g}
\end{equation}

The attention shift updates on each $j$ iteration is:

\begin{equation}
g'_{i,\, j+1} = g'_{i,\, j} \;-\; \rho \sum_{k} \frac{\partial L_k}{\partial g_i}
\end{equation}

After the settling, or stabilization as \cite{kruschke2001unified} called it, the global salience is updated via a state displacement and clamped to be non-negative:

\begin{equation}\label{eq:clamping}
\eta_{i,\, t+1} = \max\!\left[\,0,\;\; \eta_{i,t} + \alpha\!\left(g'_i - g_i\right)\right]
\end{equation}

The displacement $\left(g'_i - g_i\right)$ grows with the number of active outcomes because the summed gradient pulls $g'_{i}$ further from its initial value $g'_{i, j=1}$. When

\begin{equation}
-\alpha\!\left(g'_i - g_i\right) > \eta_{i,t}
\end{equation}

the clamp fires and $\eta_i$ resets to zero; the projection maps the value to the boundary for the next trial:

\begin{equation}
\eta_{t+1} = 0.
\end{equation}

For any model $M$ with $K$ simultaneously active outcomes, where $|K| > 1$, the magnitude of the attention update scales as $O(K)$. This is true for all models satisfying the following conditions:

\begin{enumerate}
    \item \textbf{Additive loss across outcomes}. The total loss decomposes across outcomes, $L = \sum_kL_k$, where $L_k$ is the error associated between stimulus and outcome $k$.
    \item \textbf{Gradient descent on attention}. Attention shift updates via $\Delta g = -\rho\frac{\partial L}{\partial g}$, where step-size is $\rho > 0$.
    \item \textbf{Hard non-negativity}. Attention weights are constrained to $\eta \geq 0$ via a hard projection.
    \item \textbf{Same-sign gradient reinforcement}. For multiple simultaneously active outcomes $k$, the partial derivatives $\partial L / \partial g$ for non-predicted outcomes share a sign and is not offset by the minority of the predicted outcomes.
    \item \textbf{Multi-outcome activations}. More than one outcome can be active on a single trial, $|K| > 1$.
\end{enumerate}

\begin{remark}\label{remark:same-sign}
The same-sign reinforcement condition is a property of a feature under multi-outcome learning. On a trial with $K$ co-active outcomes, let's assume that feature $i$ is predictive of some outcomes but not others. For each $k$ outcome that $i$ can predict, $\partial L_k / \partial g_i$ increases $g_i$; and for every outcome it does not predict, the shift lowers $g_i$. These descents share a sign within each group, so the summed $\sum_{k} \partial L_k / \partial g_i$ is dominated by whichever group is larger. A feature that is diagnostic for a minority of the co-active outcomes is therefore driven down by the non-predicting majority, and the magnitude of the suppressing group of outcomes grows with the number of outcomes the feature fails to inform. Once the growth exceeds the shared $\eta_{i,t}$, the clamp fires and resets it to 0. The failure is fundamentally multi-outcome: at $K = 1$ there is no majority to outvote and a dropped feature on a single-outcome trial discards a genuinely uninformative feature \citep{kruschke2001unified}. In multi-outcome learning, the mechanism strips attention from a cue in proportion to how many outcomes it is irrelevant to: it is punished for being \emph{selectively} informative.
\end{remark}

\begin{remark}\label{remark:sign-cancellation}
As a consequence of Remark \ref{remark:same-sign}, a shared salience can fail in a second, quieter way, even where the boundary is never approached. When features are predictive of some outcomes and not predictive of others, the summation-across-outcomes will have a mix of positive and negative signs, which cancel each other out. The mechanism destroys attention to a feature because it excessively punishes a feature for only being useful for some outcomes but not all. Both the settling attention shift and the salience updates are driven by the aggregated gradient $\sum_{k} \partial L_k / \partial g_i$. Suppose cue $i$ is diagnostic in opposing directions for two co-active outcomes --- its influence should be amplified for outcome $k$ but attenuated for outcome $k'$ --- so that $\partial L_k / \partial g_i$ and $\partial L_{k'} / \partial g_i$ carry opposite signs. Although each demand is individually large, $\lvert \partial L_k / \partial g_i \rvert,\; \lvert \partial L_{k'} / \partial g_i \rvert \gg 0$, their sum cancels:

\begin{equation}\label{eq:collapsing}
\sum_{k} \frac{\partial L_k}{\partial g_i} \;\approx\; 0 .
\end{equation}

The shared salience then receives no net update and $\eta_i$ is frozen, unable to acquire the outcome-specific attention the task requires because it is lost in the summation. This failure is therefore independent of the non-negativity constraint: it afflicts any rule that collapses signed, outcome-specific gradients onto a single shared state represented as a scalar.
\end{remark}

\section{Architectural Fix}\label{sec:architectural-fix}

Here, we propose outcome-indexed attention as an alternative mechanism, which removes the summation-over-outcomes from Equation \ref{eq:sum-over-k}. Individual features encode more granular information across outcomes that cannot be collapsed into a single scalar. In the event of multiple co-active outcomes, a single cue might be preferentially predictive of some while being quite uninformative for others. Similarly, the input configuration (compound cues) will determine how salient cues must be given what we are trying to predict. This means that the predictive value of a cue is not globally fixed, but locally determined -- it is outcome-driven. This outcome-indexed attention weight follows from the problem of the cue being connected to multiple outcomes simultaneously. A single (scalar) salience per stimulus is the wrong level of description that cannot account for how informative this stimulus is \emph{for this outcome} given \emph{this cue configuration}. Different cue configurations produce different predictiveness profiles across outcomes, so attention has to be indexed by outcomes, not just the cue. Thus, in this model, attention weight is indexed by the connection it modulates. Below, we implement this architectural change.

Replace the shared state $\eta_i$ with an outcome-indexed state $\eta_{ki}$, where the normalized attention will be $k$-specific:

\begin{equation}
    g_{ki} = \eta_{ki} \times s_i
\end{equation}

These attention gains are normalized through their $k$-specific vector $p$-norm:

\begin{equation}
    a_{ki} = \frac{g_{ki}}{\Big(\sum_{j} |g_{kj}|^p\Big)^{\frac{1}{p}}}
\end{equation}

Here, we can apply each outcome's gradient independently:

\begin{align}\label{eq:fix}
    \Delta g'_{ki,\, j+1} &= \;-\; \rho\, \frac{\partial L_k}{\partial g_{ki}} \\
    &= \rho s_i \|g'_{k,j}\|_p^{-1}\delta'_k\bigg(W_{ki}s_i - a'^{p-1}_{ki}o'_k\bigg)
\end{align}

The summation over $k$ is removed, which we extrapolate to the loss function:

\begin{equation}
    L_k = \frac{1}{2}(t_k - o_k)^2
\end{equation}

Each row of the attention matrix settles under its own error signal. The attention update becomes:

\begin{equation}
\eta_{ki,\, t+1} = \max\!\left[\,0,\;\; \eta_{ki,t} + \alpha\!\left(g'_{ki} - g_{ki}\right)\right]
\end{equation}

The displacement $\left(g'_{ki} - g_{ki}\right)$ is now driven by a single outcome's gradient, so its magnitude no longer scales with the number of active outcomes.

\section{Simulations}\label{sec:simulations}

We evaluate and compare the shared vector and attention matrices across three simulations procedurally increasing the input-output mapping complexity. In all simulations below, we will pretrain model weights with a delta-rule network \citep{gluck1988conditioning, rescorla1972theory} to develop non-zero and meaningful input-output representations before applying the attentional shift mechanism outlined above. All starting weights were initialized to a non-zero value by sampling from a normal distribution with a mean of 0 and standard deviation of 0.025, and the learning rate set to 0.1. The network was trained for 50 epochs, with each epoch comprising a single presentation of each stimulus. Order of presentation was randomized between each epoch. After this training, we apply the attention shift mechanism for globally shared salience vectors (Equation \ref{eq:kruschke-shift}) and outcome-indexed attention weight matrices (Equation \ref{eq:fix}); the updates are further constrained through a squashing hyperbolic tangent functions, we acquired similar results without a squashing function. For the exact equations, see Appendix \ref{appendix:derivations}. Condition 2 of Section \ref{section:problem} requires $\rho > 0$; it places no upper bound on the step-size and the collapse is claimed to be for the class of shared attention vectors, not for a particular parameterization. Thus, we sweep $\rho$ over a 0-2 range instead of fixing it across simulations. Below, we measure the stability of these mechanisms as proportion of stimulus that triggered the clamping mechanism; the boundary hit is defined as the trigger, which are all instances when attention to a feature moves below 0 as per Equation \ref{eq:kruschke-shift}. Table \ref{tab:experimental-design} shows the abstract design for the three following simulations, including the stimulus representations and the set ($T$) of teaching vectors corresponding to each stimulus. We fixed the stimulus set, $S$, throughout the simulations.

\begin{table*}[!ht]
    \centering
    \caption{Stimulus Input Patterns and Feedback Vectors Across Overlap Conditions. Rows correspond to stimulus-teacher pairs, such that stimulus A is horizontally followed by its respective teaching vector.}\label{tab:experimental-design}
    \renewcommand{\arraystretch}{1.25}
    \setlength{\tabcolsep}{3.7pt}
    \begin{tabular}{@{\hspace{2pt}}ccccc@{\hspace{8pt}}ccccc@{\hspace{8pt}}ccccc@{\hspace{8pt}}ccc}
    \toprule
    \multirow{3}{*}{Stimulus} & \multicolumn{4}{c}{Input Pattern ($S$)} & \multicolumn{13}{c}{Feedback Vectors Used in the Three Simulations ($T$)} \\
    \cmidrule(lr){2-5} \cmidrule(lr){6-18}
     & \multirow{2}{*}{$S_1$} & \multirow{2}{*}{$S_2$} & \multirow{2}{*}{$S_3$} & \multirow{2}{*}{$S_4$} & \multicolumn{5}{c}{Distinct Multi-Outcome} & \multicolumn{5}{c}{Shared Outcome Space} & \multicolumn{3}{c}{\shortstack{Distinct Singular}} \\
    \cmidrule(lr){6-10} \cmidrule(lr){11-15} \cmidrule(lr){16-18}
     & & & & & $O_1$ & $O_2$ & $O_3$ & $O_4$ & $O_5$ &  $O_1$ & $O_2$ & $O_3$ & $O_4$ & $O_5$ & $O_1$ & $O_2$ & $O_3$  \\
    \midrule
    A & 1 & 0 & 1 & 0 & 0 & 1 & 0 & 0 & 0 & 0 & 1 & 0 & 0 & 1 & 0 & 1 & 0 \\
    B & 0 & 1 & 1 & 0 & 1 & 0 & 1 & 0 & 0 & 1 & 0 & 1 & 0 & 0 & 1 & 0 & 0 \\
    C & 0 & 1 & 0 & 1 & 0 & 0 & 0 & 1 & 1 & 0 & 0 & 1 & 1 & 1 & 0 & 0 & 1 \\
    \bottomrule
    \end{tabular}
\end{table*}

\subsection{Distinctive Singular: minimal case with no conflict}\label{section:simulation-distinct}

The first instance begins with the least demanding case. The problem is defined with a disjoint outcome space of $|K| = 3$, with $|T_i| = 1$, where $T_i$ is the target set for stimulus $i$ with its teaching vector, and $|T|$ is the cardinality of $T$. Each present stimulus predicts exactly one positive outcome (receives excitation on a single output node) and faces $|K| - 1$ absent outcomes (receives no excitation), so for $|K| > 2$ the absent group strictly outnumbers the positively predictive one. Table \ref{tab:experimental-design} shows the set of $T$ used for the current simulations under Distinct Singular header. This experiment is designed to test for same-sign suppression and general boundary collapse in Remark \ref{remark:same-sign}.

\begin{figure*}[!ht]
    \centering
    \includegraphics[width=0.85\textwidth]{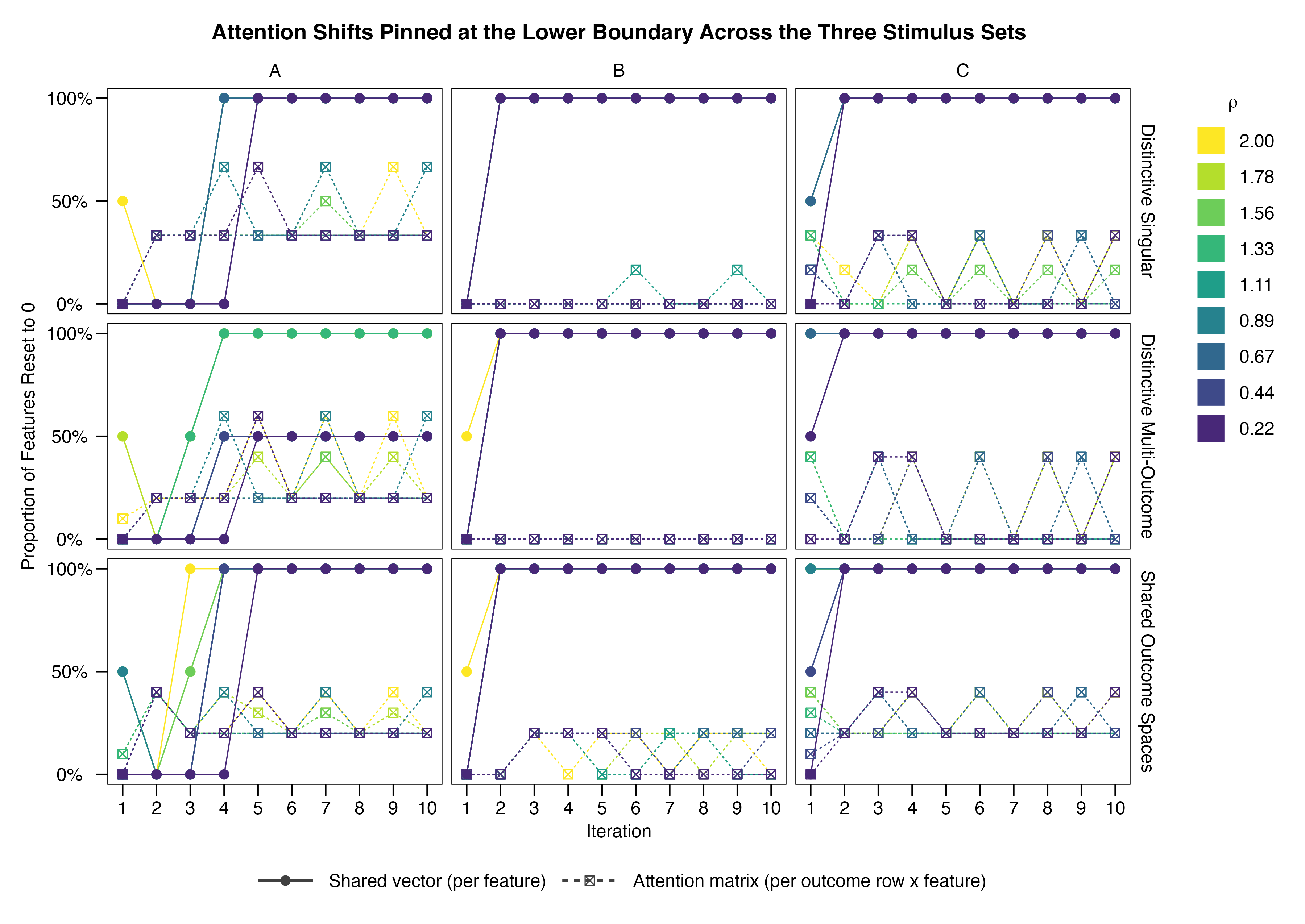}
    \caption{The boundary-hit sweeps across $\rho$: rows are the three stimulus sets (ordered from top to bottom by how much they force stimuli to compete), columns show the three stimuli, x-axis show the iteration in the attention shift mechanism, y-axis shows the proportion of eligible attention shifts reset at zero after crossing the boundary. Colour shows step-size, $\rho$, one curve per value; the attention representation is shown as shape and line-type: solid with circles for the shared vector, dashed with crossed empty squares for the outcome-indexed attention matrix.}\label{fig:boundary-hits}
\end{figure*}

\begin{figure*}[!ht]
    \centering
    \includegraphics[width=0.85\textwidth]{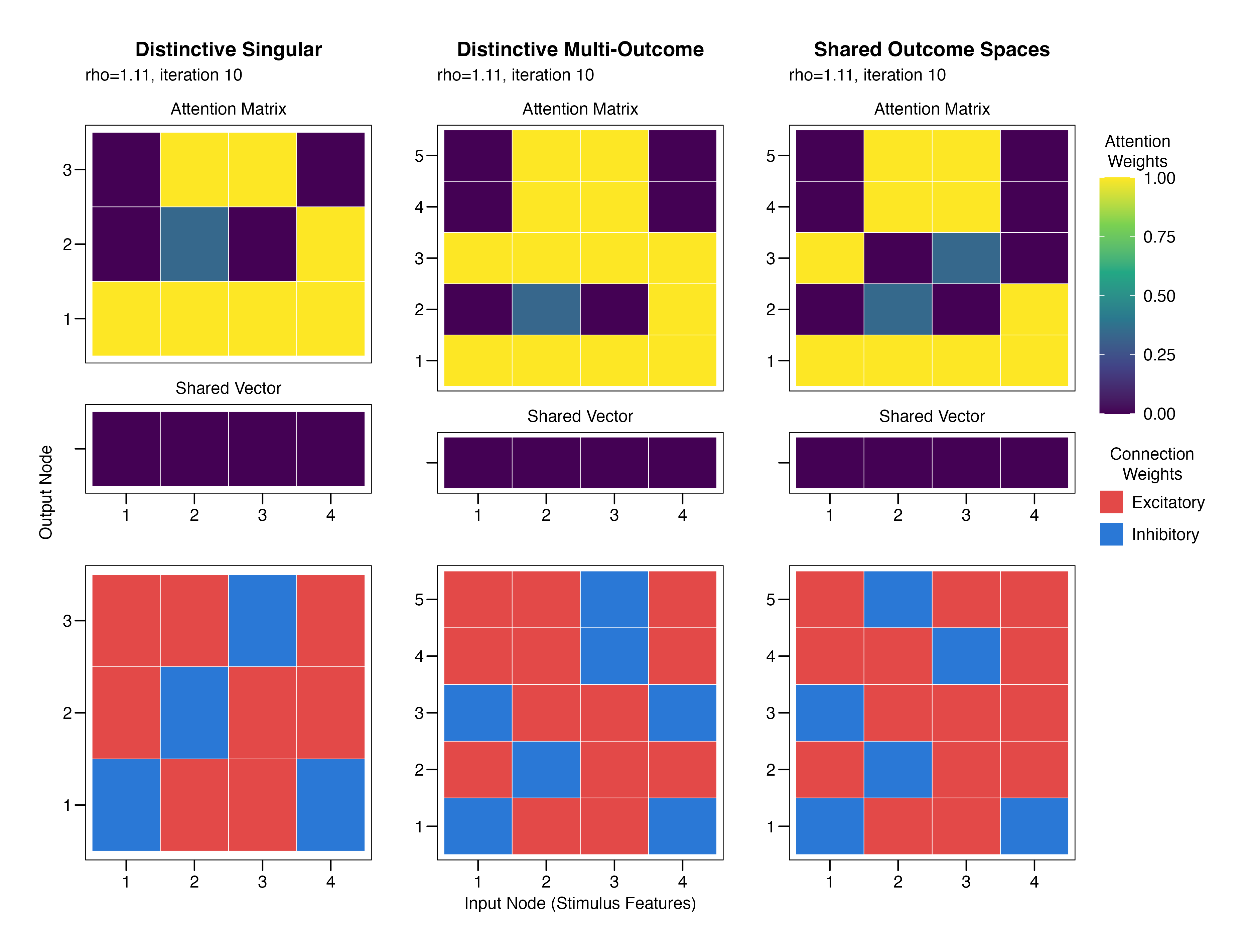}
    \caption{The three stimulus sets side by side (columns, ordered by how much they force stimuli to compete), each with the final attention gains reached after the final iteration (top row; the per-outcome attention matrix above the single shared vector) and the sign of its pretrained input-to-output connection weights (bottom row). Attention weights is shown on a continuous color scale, and normalized to a common 0-1 scale across columns, so allocations are comparable between stimulus sets. Red color indicates that the connection between input-output pairs are excitatory, blue indicates that they are inhibitory.}\label{fig:heatmaps}
\end{figure*}

The first row of Figure \ref{fig:boundary-hits} shows the proportion of features driven to the boundary for each iteration of the gradient descent, across the $\rho$ sweep. The attention shift for shared vectors consistently resets all feature-specific scalar by the end of the attention shift, and its terminal state floors all values, resulting in $g' = 0$, producing $0 - g = -g$ in the attention update; attention will decrease for all cues regardless of their informativeness. The outcome-indexed attention matrix settles on non-zero and informative values. For most iterations of the attention shift, a substantial proportion of outcome-indexed attention weights survive, resulting in an informative terminal state that encodes what the shared vector cannot: differential and selective informativeness.

The shared vector collapsed to floor value across all four features: $S_1$ and $S_4$, which are unique to a single stimulus, are indistinguishable from $S_2$ and $S_3$, which appear in two; see first column of Figure \ref{fig:heatmaps}. The vector carries no information about which features are informative because it carries no information at all. The attention weight matrix, run on the same input-output pairs, is structured: $\eta_{ki}$ is elevated for 8 out of the 12 feature-outcome mappings and is at 0 for the remainder. For example, $S_1$ is unique to stimulus $A$ and is connected to $O_2$ through excitatory and $O_1$ through inhibitory connections. Interestingly, inhibitory connections are reliably amplified through the attentional weights. Its excitatory connection to $O_3$ might seem surprising, but it is a function of the delta rule applied during learning. As the delta rule calculates error as the sum of errors across input nodes for each outcome, the positive update to $S_1 \to O_3$ is the result of the inhibitory connection developed from $S_3 \dashv O_3$, so that the output node activation matches the teaching signal of 0.

\subsection{Distinctive Multi-Outcome: no outcome overlap with multiple excited output units}\label{section:simulation-multi}

In the second instance, we increase the problem size to $|K| = 5$ while keeping the disjoint outcome space. In this problem set-up, $|T_i| \in \{1, 2\}$. The feature's demands remain same-signed for the most part, but now a majority suppression group can exist.

As before, the shared vector collapsed to floor across all four features, see middle column of Figure \ref{fig:heatmaps}. Although features present in stimulus A retains more of its informativeness relative to the previous simulations for longer and for smaller step-sizes; see second row of Figure \ref{fig:boundary-hits}. Attention matrices however encode robust representations and retain where the feature is informative.

\subsection{Shared Outcome Spaces: overlapping co-active outcomes}\label{section:simulation-shared}

In this last instance, we further increase complexity by introducing a shared outcome space with $|T_i| \in \{2, 3\}$, such that each stimulus has a single output unit that is excited and overlaps with the target teaching vector for another. Table \ref{tab:experimental-design} shows the outcome set $T$ for this simulation under Shared Outcome Space heading. $S_3$ is present in both A and B, and must support $O_2$ and $O_5$ on A trials while supporting $O_1$ and $O_3$ on B trials; $S_2$ is present in both B and C, supporting $O_1$ and $O_3$ on one and $O_3$, $O_4$ and $O_5$ on the other. $S_2$ and $S_3$ each carry demands of opposite sign within a trial; thus opposite signs coexist within a trial. This is the silent failure mode described in Remark \ref{remark:sign-cancellation}. The shared salience for these features receives no net (or infinitesimal) updates, while the attention matrix stores the opposing demands on outcome-indexed rows; see Figure \ref{fig:heatmaps}.

\section{Summary}\label{sec:summary}

Here we presented the conditions under which attentional learning fails in feed-forward network models of learning, and proposed \emph{outcome-driven attentional learning} as a solution to the breakdown. This result has implications for a range of models using attention shift as an update mechanism, such as EXIT \citep{kruschke2001unified}, RASHNL \citep{kruschke1999model}, and their neural network derivatives \citep{paskewitz2020dissecting}. The three simulations we ran established that: the collapse is not the by-product of step-size as it occurs across all explored range of $\rho > 0$; the sign-cancellation failure described in Remark \ref{remark:sign-cancellation} is independent of the non-negativity constrains; lastly, our proposed model architecture (outcome-indexed attention weight matrices) are a sufficient solution -- decoupling gradients by outcomes removes the failure modes without altering the underlying learning rule. The fix reduces to one structural change of

\begin{align}
\sum_{k} \frac{\partial L_k}{\partial g_j} \quad\longrightarrow\quad \frac{\partial L_k}{\partial g_{kj}}
\end{align}

applied independently per outcome $k$ for the attention matrix $\boldsymbol{\eta} \in \mathbb{R}_{\geq 0}^{K \times J}$. This solution produces several theoretically interesting consequences. The outcome-indexed attention weights enable features to take on outcome-specific importance, so that different features are selectively activated depending on what the system is attempting to predict. Attentional reallocation within each trial is determined by what the system is trying to do rather than cue-specific associability independent of the system's overall purpose. Features can be diagnostic of some outcomes, but not others, and this information is used depending on what available outcomes are excitable.

Across our simulations in Section \ref{section:simulation-shared} and \ref{section:simulation-multi}, outcomes are shared between inputs. These scenarios have strong implications for extending similar attentional processes to reinforcement learning environment with probabilistic feature-outcome mappings \citep{jones2010integrating, canas2010attention}. In this case, features are connected to more than one output nodes via excitatory connections, which will cause scalar salience to break down. The attention matrix is structurally incapable of having that instability.

We also observed an interesting mapping between attention and connection weights. In a matrix-like representation, dimensional attention vectors encode information about both excitatory and inhibitory connections separately, which are selectively amplified by the attentional gating procedure. In almost all simulations, inhibitory connections were always amplified, but excitatory connection were less likely to increase in salience. This configuration was sufficient for the model to learn the current input-output mappings, but it invokes interesting implications for early phases of learning. Features with close-to-zero initial weights become predictive of absence of an outcome early on and acquire strong attentional weight. Excitatory connections to outcomes then develop undisturbed with standard excitatory connection that do not need attentional augmentation. However, initializing connection weights close to zero is a choice, and principled alternative approaches exist \citep{spicer2021representing,rumelhart1986parallel}.

We do not present evidence that prior models are inadequate. On the contrary, we kept their core computational principles (attention adjusted on gradient descent on error) intact and extended it to more challenging environment. \cite{mackintosh1975theory} formalizes salience, also called associability, as the property of a feature that is acquired through its history of predictive success. Our results do not contradict this position, as our attention matrix architecture becomes indistinguishable from the Mackintosh tradition under $K = 1$.

In psychology, we often stabilize our models by reducing the complexity of the experiments \citep[``externally restricting the input environment'';][]{grossberg1976badaptive, grossberg1976aadaptive}. The problem than becomes one of model capacity and not adequacy. The increase in environmental complexity presented here makes cognitive models more ecologically valid, which in turn improves the mapping between the model, the experimental paradigms, and the real-world conditions. In the current case, improved model capacity allows the computational mechanism to generalize to multiple co-active $K$ outcomes. In practice, we often predict more than a single event. If we diagnose a disease early, we predict not just the disease label, but also symptoms that are yet to be experienced, what potential treatment might be used given hospital resources or symptom combinations, and predict costs for the patient. In many instances, the number of outcomes we anticipate exceeds the number of features presently observed.

\section{Conclusion}\label{sec:conclusion}

Selective attention implemented as a globally shared salience vector is unstable under multi-outcome learning. Its instability emerges from its architecture, persists across a range of step-sizes, and remains invisible under single-outcome paradigms which these models were designed to accommodate. We identified the instability to result from summing outcome-indexed gradients onto a single shared state. Any architecture implementing this summation is within scope. Indexing attention by outcome restores stable learning with one relatively small architectural change. Outcome-indexed attention weights also encodes more granular representations that scalar saliences cannot: informativeness of a feature is specific to the outcome and the configuration in which it appears. The architecture change proposed here remains minimal, and it is a precondition for extending attentional learning for richer and more demanding environments in which real-world predictions usually take place.

\section*{Acknowledgements}

I would like to thank Xin Sui for helpful discussions on how to approach gradient descent. I would like to further thank Maciek Szul and Ehsan Kakaei for helpful comments on the manuscript.

\section*{Open Science}

All simulation code is available on GitHub: \url{https://github.com/lenarddome/tue010-attention-unstability}.

\bibliographystyle{plainnat}
\bibliography{bibliography}

\appendix

\section{Attention shift equations}\label{appendix:derivations}

Attention shift for globally shared salience vectors are defined as:

\begin{align}
\Delta g'_{i} &= \tanh\left(-\rho \frac{\partial E}{\partial g_{i}}\right) \\
&= \tanh\left[\rho s_i \|g'\|_p^{-1}\sum_k(W_{ki}s_i - a'^{p-1}_{i}o'_k)\delta'_k\right]
\end{align}

where $\rho$ is a positive constant, denoting the step size for the gradient descent, called the attention shift rate; $\tanh$ is a squashing hyperbolic tangent function that we apply to further constrain updates to lie between -1 and 1.

Attention shift for matrix-representation for outcome-indexed attention weights are defined as:

\begin{align}
\Delta g'_{ki} &= \tanh\left(-\rho \frac{\partial E_k}{\partial g_{ki}}\right) \\
&= \tanh\left[\rho s_i \|g'_k\|_p^{-1}\delta'_k\big(W_{ki}s_i - a'^{p-1}_{ki}o'_k\big)\right]
\end{align}

\end{document}